\pdfoutput=1
\documentclass[12pt]{article}
\usepackage{amsmath, amssymb, amsthm, bm, fullpage}
\usepackage{xcolor}
\usepackage{booktabs}
\usepackage{graphicx}
\usepackage[hang,flushmargin]{footmisc}

\usepackage[round]{natbib}

\title{Robustness of Refugee-Matching Gains to Off-Policy Evaluation Choices}
\author{Kirk Bansak,\textsuperscript{a,b} 
           Elisabeth Paulson,\textsuperscript{a,c} 
           Dominik Rothenhäusler,\textsuperscript{d} \\
           Jeremy Ferwerda,\textsuperscript{a,e} 
           Jens Hainmueller,\textsuperscript{a,f}
           Michael Hotard\textsuperscript{a} 
         }

\date{\small \textsuperscript{a}Immigration Policy Lab, Stanford University \\
\textsuperscript{b}Department of Political Science, University of California, Berkeley \\
\textsuperscript{c}Technology and Operations Management Unit, Harvard Business School  \\
\textsuperscript{d}Department of Statistics, Stanford University \\
\textsuperscript{e}Department of Government, Dartmouth College \\
\textsuperscript{f}Department of Political Science, Stanford University \\ \vspace{0.4cm} \large
\today
}

\begin{document}
\maketitle

\begin{abstract}
\noindent Previous research has investigated the potential of refugee matching for boosting refugee outcomes, first considered by \cite{bansak2018improving}. This paper demonstrates the stability of counterfactual impact evaluation results in the context of refugee matching in the United States using a range of off-policy evaluation methods. In order to estimate counterfactual impact and test the robustness of our results, we employ several evaluation methods, including inverse probability weighting (IPW) and multiple variants of augmented inverse probability weighting (AIPW). We also consider various modifications, including alternative modeling architectures and different assignment procedures. The impact estimates remain consistent in magnitude in all scenarios as well as statistically significant in most cases. Furthermore, the estimates are also consistent with the results originally presented in \cite{bansak2018improving}.
\end{abstract}

\section{Introduction}

The idea of algorithmic refugee assignment to improve refugee outcomes within their host countries was originally proposed by \cite{bansak2018improving}, which also presented preliminary counterfactual impact evaluations. Since then, various studies have considered a range of extensions \cite[e.g][]{golz2019migration,ahani2021placement,acharya2020matching,freund2023group,ahani2024dynamic,bansak2024outcome,bansak2024learning,jain2025ctrl,rodriguez2025dual,bansak2026dynamic}. This paper extends the impact evaluations in \cite{bansak2018improving} of the potential offered by algorithmic refugee assignment in the United States, demonstrating the stability of the results using a range of off-policy evaluation methods. 

In their original analyses, \cite{bansak2018improving} employed common model-based policy evaluation procedures. A number of studies have discussed issues with model-based counterfactual estimation and policy evaluation, such as the possibility of ``winner's curse" bias \cite[e.g.][]{andrews2024inference,zrnic2025flexible,bastani2026winner}. For example, \cite{bastani2026winner} show that the winner's curse bias can be large on a synthetic dataset.

In this paper, we employ more robust evaluation methods to estimate the counterfactual impact of refugee matching in the United States, including inverse probability weighting (IPW) and multiple variants of augmented inverse probability weighting (AIPW). We also consider various modifications, including alternative modeling architectures and different assignment procedures. In all scenarios, the impact estimates remain consistent in magnitude and statistically significant in most cases. Furthermore, the estimates are also consistent with the results originally presented in \cite{bansak2018improving}. As one of our evaluation scenarios we use the exact same data and models originally used in \cite{bansak2018improving}, and our robust evaluation methods yield similar impact estimates as the model-based methodology originally used in \cite{bansak2018improving}, demonstrating that those original results were not driven by winner's curse bias.

In sum, we find stable impact evaluation results using a range of robust evaluation methods. The results underscore the potential of data-driven refugee matching to meaningfully improve employment outcomes.

\section{Setup}

We observe a finite population of individuals in a test/evaluation set indexed by $i=1,\dots,N$. Each individual was historically assigned to one of $K$ locations, denoted $A_i \in \{1,\dots,K\}$, and experienced employment outcome $Y_i \in \{0,1\}$. A proposed algorithmic matching assigns each individual to a new location $g_i \in \{1,\dots,K\}$. We wish to estimate the average employment rate that would have been obtained if the algorithmic assignment $g = (g_1,\dots,g_N)$ had been used instead of the historical assignment $A$.

We use data on historical refugee assignments and outcomes from one of the largest resettlement agencies in the United States. Historical assignment was quasi-random.\footnote{Based on empirical assessments and conversations with representatives from the resettlement agency, we know that the historical assignments of free cases were done in a manner unrelated to expected outcomes, though some case officers may have used covariates that we also have access to. Therefore, the historical assignment satisfies ignorability.} Therefore, we consider two possibilities. The first is that local-level assignment probabilities $\pi_{a,i}$, denoting the probability that an individual $i$ was assigned to location $a$, are homogeneous across individuals such that $\pi_{a,i} = \pi(a)$, which is known based upon the historical empirical counts. Second, as a robustness check, we also consider the possibility that $\pi_{a,i}$ are conditional on covariates, $\pi_{a,i} = \pi(a | X_i)$, in which case we estimate those propensities. We assume no interference (SUTVA), and we let $Y_i(a)$ denote the potential employment outcome for individual $i$ under location $a$. We observe $Y_i = Y_i(A_i)$.

We are interested in estimating the policy value
\[
V(g) = \frac{1}{N} \sum_{i=1}^N Y_i(g_i).
\]
To predict counterfactual rewards used in the algorithm, we train supervised machine learning models on a set of training data (separate from the evaluation set). Let $\mu_i(a)$ denote an estimate of the conditional expected outcome
$\mathbb{E}[Y_i(a)|X_i]$. For each individual, define
\[
\mu_{A,i} = \mu_i(A_i), \qquad 
\mu_{g,i} = \mu_i(g_i), \qquad
\pi_{A,i} = \pi_{A_i,i}
\]
where, as noted above, we consider both possibilities that $\pi_{A,i} = \pi(A_i)$ and $\pi_{A,i} = \pi(A_i | X_i)$.

We perform counterfactual algorithmic assignment on two data setups. The first is the exact data used in \cite{bansak2018improving}, in which case we form our evaluation set of assignments by pooling their evaluation sets from 2015 Q4 to 2016 Q3, and we use their exact estimates of $\mathbb{E}[Y_i(a)]$, which were obtained from a predictive model where prior data are used to train Stochastic Gradient Boosted Tree (SGBT) ensembles. The exact models, predictions, and assignments used in \cite{bansak2018improving} are used here, constituting a replication of \cite{bansak2018improving}.

The second data setup uses updated data, where our evaluation set of assignments includes all historical assignments made by our resettlement partner over the entire year of 2016. In this case, estimates of $\mathbb{E}[Y_i(a)]$ are obtained from a predictive model where prior data are used to train Bayesian Additive Regression Tree (BART) models. Furthermore in this setup, while we employ location-specific BART models for locations that are sufficiently large, we use a pooled BART model for small locations (with location-specific indicators as predictors). Using these predictions, we formulate new counterfactual assignments for our evaluation set. In doing so, we consider both offline and online assignments, where for the latter we implement the online assignment algorithm presented in \cite{bansak2026dynamic}.

In both data setups, when we consider the possibility that $\pi_{A,i} = \pi(A_i)$, we use the historical empirical counts to calculate the known propensities (i.e. ``empirical propensities"). When we consider the possibility that $\pi_{A,i} = \pi(A_i | X_i)$, we estimate the propensities using random forest probability models (i.e. ``estimated propensities"). In addition, to address the impact of small propensity scores on the variance of our estimators (described below), we also consider a modification to our algorithmic design that pools small locations (those with $\pi(a) < 0.01$) into a single pseudo-location. When using this modification, the pooling applies to both the matching and evaluation procedure. Effectively, this says that anyone assigned to a ``small location'' is randomly (proportionally) assigned to one of the locations in the pseudo-location pool. This strategy can theoretically introduce a downward bias in the point estimate but should also reduce variance. It is also a valid procedure to use in practice. 

In both data setups, we make the counterfactual assignments on the level of cases (i.e. families) and evaluate impact on the basis of individuals, just as in \cite{bansak2018improving}.

\section{Estimators}

We estimate the counterfactual impact of algorithmic assignment via three different estimators, described below.

\subsection{Inverse Probability Weighting (IPW)}

The IPW estimator uses only the randomized historical assignment and does \emph{not} use model predictions. 
As noted above, let $\pi_{a,i}$ denote the historical probability of individual $i$ being assigned to location $a$. Define the indicator $\mathbf{1}(A_i = g_i)$, which equals one if individual $i$ happens to have been assigned historically to the same location as the algorithmic policy recommends. The IPW estimator is
\begin{equation}
  \widehat V_{\mathrm{IPW}}
  =
  \frac{
     \sum_{i=1}^N \mathbf{1}(A_i = g_i)\,\dfrac{Y_i}{\pi_{A,i}}
   }{
     \sum_{i=1}^N \mathbf{1}(A_i = g_i)\,\dfrac{1}{\pi_{A,i}}
   }.
\end{equation}

Identification of IPW requires the following conditions:

\begin{enumerate}
\item Ignorability:
      $Y_i(a) \perp A_i \:\: | \:\: X_i$ for all $a$.
\item Positivity: If the policy assigns some individuals to 
      $g_i = a$, then $\pi_{a,i} > 0$. 
\item SUTVA: No interference across individuals.
\end{enumerate}

IPW uses only observed outcomes $Y_i$ and does not involve any predictive model. However, the possibility of small propensity scores is known to sometimes ``blow up" the variance of the IPW estimator. To address this, we also consider the pooling modification described earlier, whereby small locations (those with $\pi(a)<0.01$) are pooled into a single pseudo-location.

\subsection{Augmented IPW (AIPW)}

AIPW combines inverse probability weighting with an outcome regression model. The AIPW estimator is
\begin{equation}
  \widehat V_{\mathrm{AIPW}}
  =
  \frac{1}{N}
  \sum_{i=1}^N
    \left[
      \mu_{g,i}
      +
      \mathbf{1}(A_i = g_i)\,\frac{Y_i - \mu_{A,i}}{\pi_{A,i}}
    \right].
\end{equation}

AIPW is doubly robust. It is consistent if either:
\begin{enumerate}
\item The historical assignment probabilities $\pi_{A,i}$ are correctly specified, or
\item The outcome regression model $\mu_i(a)$ is correctly specified.
\end{enumerate}
AIPW also offers greater statistical efficiency relative to IPW. For AIPW, we also consider both the standard matching procedure and pooled matching procedure described above.

\subsection{AIPW-local (AIPWl)} 

We also consider a variant of AIPW that we call AIPW-local, or AIPWl. For the AIPWl estimator, the augmentation term replaces the marginal propensity $\pi_{A,i}$ with a location-specific estimate of
\[
\pi_{\mathrm{L}}(a)
\;=\;
\Pr(A_i = a \mid g_i = a)
\;\approx\;
\frac{
  \#\{ i : A_i = g_i = a \}
}{
  \#\{ i : g_i = a \}
}.
\]
Define
\[
\psi_i^{\mathrm{L}}
=
\mu_{g,i}
+
\mathbf{1}(A_i = g_i)\,
\frac{Y_i - \mu_{A,i}}{\pi_{\mathrm{L}}(g_i)}.
\]
The AIPWl estimator is
\[
\widehat V_{\mathrm{AIPWl}}
=
\frac{1}{N}\sum_{i=1}^N \psi_i^{\mathrm{L}}.
\]

\subsection{Model-Based}

As a benchmark comparison, we also consider the model-based evaluation approach employed in \cite{bansak2018improving} that calculates counterfactual employment and gains solely by using the estimates of $\mathbb{E}[Y_i(a)]$ that are also used for the assignment decisions. In the Results section later, we refer to the results using this method simply as ``model-based."

\section{Design-based Uncertainty Quantification}

We are interested in estimating, for a given population and a given matching/assignment, the overall employment rate that this population and assignment would result in. Therefore, we treat the population, historical refugee arrivals that were used for training $\mu$, and policy assignment $g$ as fixed. We use the quasi-randomness of the historical assignments used for evaluation for uncertainty quantification. That is, we use variation of $A_i$, $i=1,\ldots,N$. In the following, we describe uncertainty quantification for the quasi-random case ($\pi_{A,i} = \pi(A_i)$).

\subsection{Variance of AIPW and AIPWl}
\begin{equation*}
   V- \hat V_{\text{AIPW}}  = \frac{1}{N} \sum_{i=1}^N Y_i(g_i) - \mu_{g,i} - 1_{A_i = g_i} \frac{Y_i - \mu_{A,i}}{\pi_{A,i}} = \frac{1}{N} \sum_{i=1}^N ( \pi(g_i) - 1_{A_i = g_i}) \frac{(Y_i(g_i) - \mu_{g,i})}{\pi(g_i)}.
\end{equation*}
Then,
\begin{equation*}
    \text{Var}( V-\hat V_{\text{AIPW}}) = \frac{1}{N^2} \sum_{i=1}^N \pi(g_i) (1 - \pi(g_i))  \left( \frac{(Y_i(g_i) - \mu_{g,i})}{\pi(g_i)} \right)^2
\end{equation*}
We cannot directly compute this variance since we do not observe all terms for everyone. An unbiased estimate is
\begin{align*}
    \widehat{\text{Var}}(\hat V_{\text{AIPW}} - V) &= \frac{1}{N^2} \sum_{i=1}^N 1_{A_i = g_i} (1 - \pi(g_i)) \left( \frac{(Y_i(g_i) - \mu_{g,i})}{\pi(g_i)} \right)^2 \\
    &= \frac{1}{N^2} \sum_{i : A_i = g_i} (1 - \pi_{A,i}) \left( \frac{(Y_i - \mu_{A,i})}{\pi_{A,i}} \right)^2
\end{align*}
The same approximate variance formula applies also to AIPWl, simply replacing $\pi(A_i)$ with $\pi_D(A_i)$.

\subsection{Variance of IPW}

First,
\begin{equation*} 
 \hat{V}_\text{IPW} - V = \frac{\frac{1}{N} \sum_{i=1}^N \frac{1_{A_i = g_i} }{\pi_{A,i}} (Y_i - V )}{\frac{1}{N} \sum_{i=1}^N \frac{1_{A_i = g_i}}{\pi_{A,i}}  } = \frac{1}{N} \sum_{i=1}^N  \frac{1_{A_i = g_i}}{\pi_{A,i}} (Y_i(g_i) - V )  + o_P(1/\sqrt{N})
\end{equation*}
where we used that the numerator is $O_P(1/\sqrt{N})$ and the denominator converges to $1$ (since potential outcomes are bounded). Then the variance of the right hand side (ignoring lower-order terms) is
\begin{equation*}
   \frac{1}{N^2} \sum_{i=1}^N \pi(g_i) (1- \pi(g_i)) \left( \frac{Y_i(g_i) - V}{\pi(g_i)} \right)^2,
\end{equation*}
which can be estimated via 
\begin{equation*}
 \widehat{\text{Var}}( V- \hat V_{\text{IPW}})  = 
   \frac{1}{N^2} \sum_{i: A_i = g_i}  (1- \pi_{A,i}) \left( \frac{Y_i - \hat V_\text{IPW}}{\pi_{A,i}} \right)^2.
\end{equation*}

\section{Results}

The results across the various setups described above are shown in Tables \ref{tab:set1_offline_empprop_nopool}--\ref{tab:set2_online_estprop_nopool} in the Appendix. Each table includes point estimates for the employment rate (policy value) under algorithmic assignment according to each evaluation method. Also displayed are the point estimates, variance estimates, and 95\% confidence intervals for the projected employment gains from algorithmic assignment relative to the status quo, in terms of percentage-point gains. As shown in the tables, all of the evaluation methods yield generally comparable estimates across the board. In addition, the benchmark model-based method does not appear to be overly or consistently optimistic. Figures \ref{fig:ppgains} and \ref{fig:percgains} below present a visual summary of the results in terms of estimated gains. Figure \ref{fig:ppgains} displays percentage-point gains relative to the status quo baseline, while Figure \ref{fig:percgains} displays percent gains over the baseline.

\begin{figure}[ht!]
    \centering
    \includegraphics[width=0.85\linewidth]{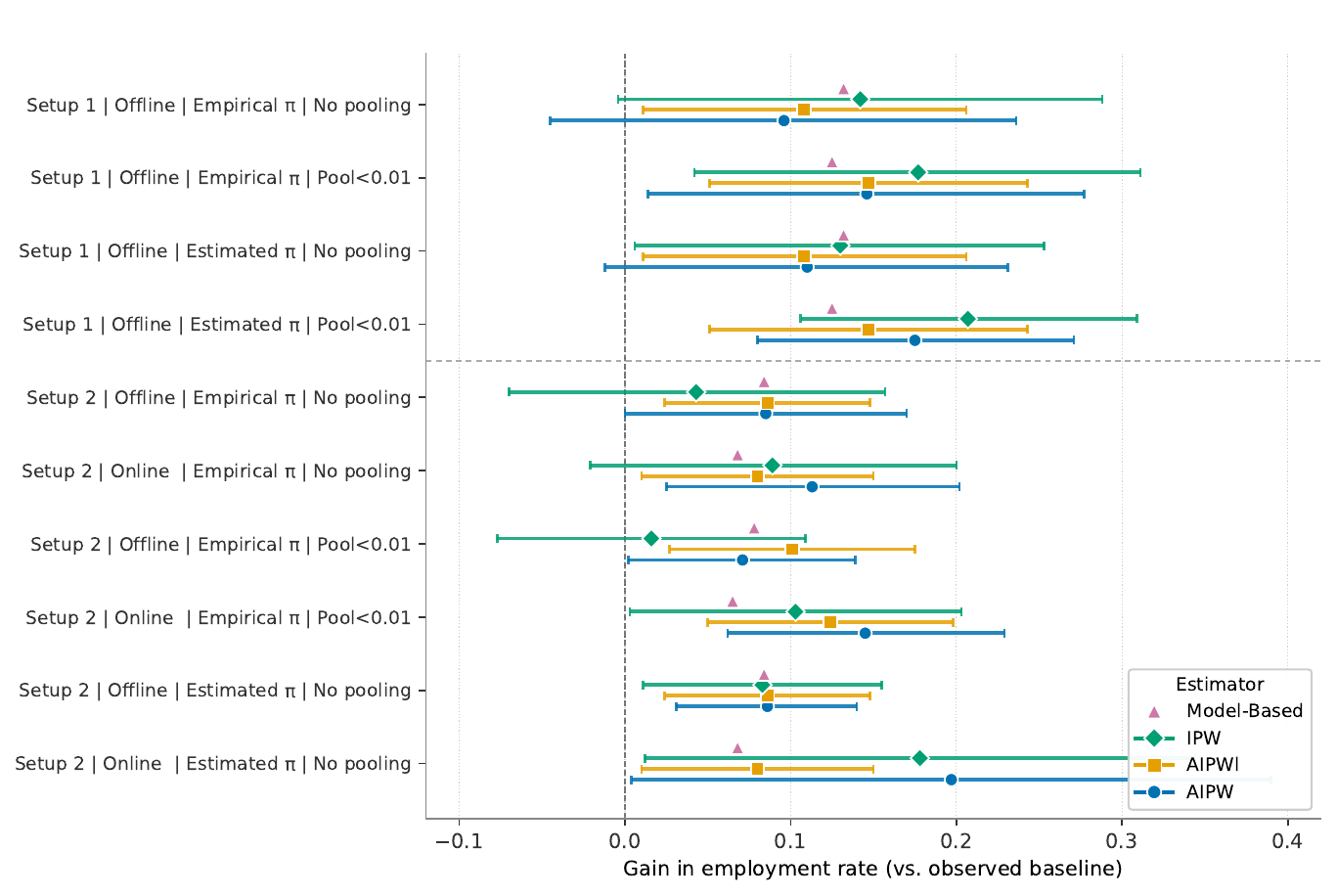}
    \caption{Percentage-Point Gains Above the Observed Baseline}
    \label{fig:ppgains}
\end{figure}

$\:$ \\ \\

\begin{figure}[ht!]
    \centering
    \includegraphics[width=0.85\linewidth]{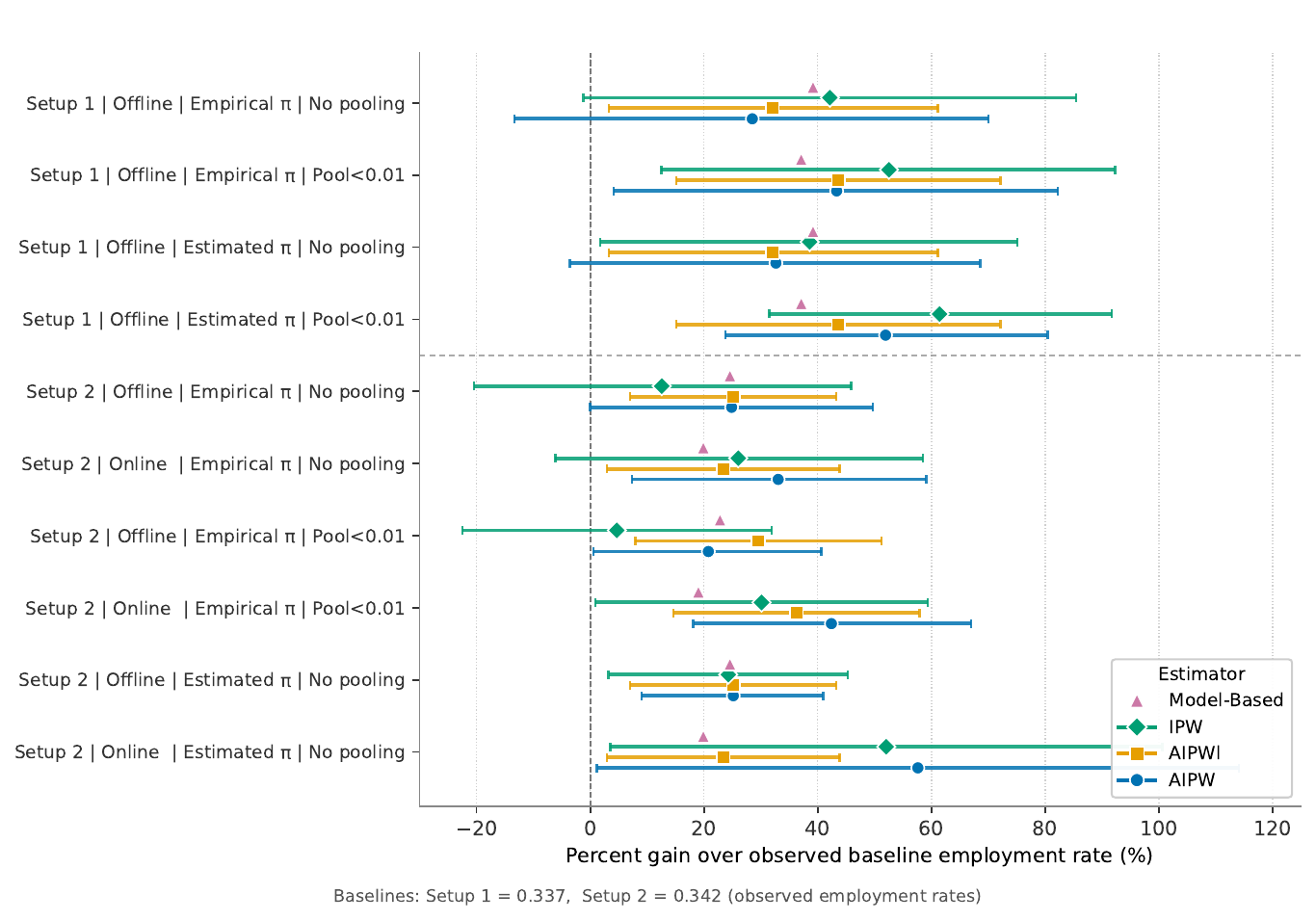}
    \caption{Percent Gains Over the Observed Baseline}
    \label{fig:percgains}
\end{figure}

\clearpage

\section{Discussion}

The results, which substantively hold across a range of counterfactual evaluation methods, underscore the potential of data-driven refugee matching to meaningfully improve employment outcomes. In addition, as can be clearly seen in our evaluations, the results produced by the benchmark model-based method---including the results originally reported in \cite{bansak2018improving}---are not overly optimistic. That is, we find no evidence that the model-based estimated gains are an artifact of winner's curse bias. This may be attributable to the specific ML methods and modeling we employed, specifically our SGBT and BART models. 

SGBT models have several forms of regularization built in, including base learner (i.e. tree) shrinkage, constraints on tree size, and training on random subsamples of the data. If such regularization led to predictive flattening (or mean reversion) across the models, that could have led to systematic under-prediction of rewards at the upper end of the distribution of reward estimates.

With respect to BART, Bayesian methods are naturally less vulnerable to winner’s curse bias because of their Bayesian structure \citep{efron2011tweedie,ferguson2013empirical,andrews2024inference}. More specifically, posterior means are Bayes-unbiased\footnote{Bayes-unbiased is similar to unbiasedness, but unbiased would be the wrong word here; unbiased means that the parameter is fixed and we average over uncertainty of the data. In the Bayesian world, the data is fixed and we average over the uncertainty in the parameters.} (i.e., they hit the target on average, conditional on all observed data). Therefore, if we make decisions based on the data (such as assigning someone), then summing the corresponding posterior means is also Bayes-unbiased for the counterfactual reward. Of course, the reliability of these estimates depends on whether the prior is reasonable. Furthermore, in our Data and Modeling Setup 2, we employ location-specific BART models for locations that are sufficiently large, but we use a pooled BART model for small locations (with location-specific indicators as predictors). Therefore, in general the predictions are likely being pushed towards the mean.

In sum, if our ML models pick up on \textit{enough} interactions (i.e. heterogeneity across locations) to find a good assignment, but they still \textit{under-predict} those interactions, it is quite possible that the true reward of that assignment is actually larger than what the models themselves would predict. The adjustments of IPW and AIPW correct for this.

\section*{Acknowledgments}

The authors thank Global Refuge for access to data and guidance. The data used in this study were provided under a collaboration research agreement with Global Refuge. This work is associated with the GeoMatch project within the Immigration Policy Lab (IPL) at Stanford University and ETH Zurich. The GeoMatch project is supported by funding from the Charles Koch Foundation, Google.org, Open Society Foundations, and Stanford Impact Labs. Rothenh\"ausler gratefully acknowledges support as a David Huntington Faculty Scholar, Chamber Fellow, and from the Dieter Schwarz Foundation.

\clearpage

\bibliographystyle{apalike}
\bibliography{refs}

\clearpage

\appendix
\setcounter{table}{0}
\renewcommand{\thetable}{A\arabic{table}}

\section{Appendix}

\subsection{Results from Data and Modeling Setup 1}

\small This Data and Modeling Setup 1 constitutes the original setup and assignments from \cite{bansak2018improving}. Note that the Gains below are relative to the empirically observed employment rate of $0.337$. Confidence intervals (CI) are 95\%.

\begin{table}[ht!]
\caption{Using offline assignment and empirical propensity scores. No pooling.} \label{tab:set1_offline_empprop_nopool}
\centering \small
\begin{tabular}[t]{lllll}
\toprule
  & AIPW & AIPWl & IPW & Model-Based\\
\midrule
Point Estimate & 0.432 & 0.445 & 0.479 & 0.469\\
Gains & 0.096 & 0.108 & 0.142 & 0.132\\
Var(Gains) & (0.0052) & (0.0025) & (0.0055) & NA\\
CI of Gains & {}[-0.045, 0.236] & {}[0.011, 0.206] & {}[-0.004, 0.288] & NA\\
\bottomrule
\end{tabular}
\end{table}

\vspace{-0.3cm}

\begin{table}[ht!]
\caption{Using offline assignment and empirical propensity scores. Pooling \% = 0.01}
\centering \small
\begin{tabular}[t]{lllll}
\toprule
  & AIPW & AIPWl & IPW & Model-Based\\
\midrule
Point Estimate & 0.483 & 0.483 & 0.514 & 0.462\\
Gains & 0.146 & 0.147 & 0.177 & 0.125\\
Var(Gains) & (0.0045) & (0.0024) & (0.0047) & NA\\
CI of Gains & {}[0.014, 0.277] & {}[0.051, 0.243] & {}[0.042, 0.311] & NA\\
\bottomrule
\end{tabular}
\end{table}

\vspace{-0.3cm}

\begin{table}[ht!]
\caption{Using offline assignment and estimated propensity scores. No pooling.}
\centering \small
\begin{tabular}[t]{lllll}
\toprule
  & AIPW & AIPWl & IPW & Model-Based\\
\midrule
Point Estimate & 0.447 & 0.445 & 0.467 & 0.469\\
Gains & 0.11 & 0.108 & 0.13 & 0.132\\
Var(Gains) & (0.0038) & (0.0025) & (0.004) & NA\\
CI of Gains & {}[-0.012, 0.231] & {}[0.011, 0.206] & {}[0.006, 0.253] & NA\\
\bottomrule
\end{tabular}
\end{table}

\vspace{-0.3cm}

\begin{table}[ht!]
\caption{Using offline assignment and estimated propensity scores. Pooling \% = 0.01.}
\centering \small
\begin{tabular}[t]{lllll}
\toprule
  & AIPW & AIPWl & IPW & Model-Based\\
\midrule
Point Estimate & 0.512 & 0.483 & 0.544 & 0.462\\
Gains & 0.175 & 0.147 & 0.207 & 0.125\\
Var(Gains) & (0.0024) & (0.0024) & (0.0027) & NA\\
CI of Gains & {}[0.08, 0.271] & {}[0.051, 0.243] & {}[0.106, 0.309] & NA\\
\bottomrule
\end{tabular}
\end{table}

\clearpage

\subsection{Results from Data and Modeling Setup 2}

Note that the Gains below are relative to the empirically observed employment rate of $0.342$. Confidence intervals (CI) are 95\%.

\begin{table}[ht!]
\caption{Using offline assignment and empirical propensity scores. No pooling.}
\centering \small
\begin{tabular}[t]{lllll}
\toprule
  & AIPW & AIPWl & IPW & Model-Based\\
\midrule
Point Estimate & 0.427 & 0.427 & 0.385 & 0.425\\
Gains & 0.085 & 0.086 & 0.043 & 0.084\\
Var(Gains) & (0.0019) & (0.001) & (0.0033) & NA \\
CI of Gains & $[0.000, \; 0.170]$ & $[0.024, \; 0.148]$ & $[-0.070, \; 0.157]$ & NA \\
\bottomrule
\end{tabular}
\end{table}

\begin{table}[ht!]
\caption{Using online assignment and empirical propensity scores. No pooling.}
\centering \small
\begin{tabular}[t]{lllll}
\toprule
  & AIPW & AIPWl & IPW & Model-Based\\
\midrule
Point Estimate & 0.455 & 0.422 & 0.431 & 0.41\\
Gains & 0.113 & 0.08 & 0.089 & 0.068\\
Var(Gains) & (0.002) & (0.0013) & (0.0032) & NA \\
CI of Gains & $[0.025, \; 0.202]$ & $[0.010, \; 0.150]$ & $[-0.021, \; 0.200]$ & NA \\
\bottomrule
\end{tabular}
\end{table}

\begin{table}[ht!]
\caption{Using offline assignment and empirical propensity scores. Pooling \% = 0.01.}
\centering \small
\begin{tabular}[t]{lllll}
\toprule
  & AIPW & AIPWl & IPW & Model-Based\\
\midrule
Point Estimate & 0.412 & 0.442 & 0.357 & 0.419\\
Gains & 0.071 & 0.101 & 0.016 & 0.078\\
Var(Gains) & (0.0012) & (0.0014) & (0.0023) & NA \\
CI of Gains & $[0.002, \; 0.139]$ & $[0.027, \; 0.175]$ & $[-0.077, \; 0.109]$ & NA \\
\bottomrule
\end{tabular}
\end{table}

\clearpage

\begin{table}[ht!]
\caption{Using online assignment and empirical propensity scores. Pooling \% = 0.01.}
\centering \small
\begin{tabular}[t]{lllll}
\toprule
  & AIPW & AIPWl & IPW & Model-Based\\
\midrule
Point Estimate & 0.487 & 0.465 & 0.444 & 0.406\\
Gains & 0.145 & 0.124 & 0.103 & 0.065\\
Var(Gains) & (0.0018) & (0.0014) & (0.0026) & NA \\
CI of Gains & $[0.062, \; 0.229]$ & $[0.050, \; 0.198]$ & $[0.003, \; 0.203]$ & NA \\
\bottomrule
\end{tabular}
\end{table}

\begin{table}[ht!]
\caption{Using offline assignment and estimated propensity scores. No pooling.}
\centering \small
\begin{tabular}[t]{lllll}
\toprule
  & AIPW & AIPWl & IPW & Model-Based\\
\midrule
Point Estimate & 0.427 & 0.427 & 0.424 & 0.425\\
Gains & 0.086 & 0.086 & 0.083 & 0.084\\
Var(Gains) & (0.0008) & (0.001) & (0.0013) & NA \\
CI of Gains & $[0.031, \; 0.140]$ & $[0.024, \; 0.148]$ & $[0.011, \; 0.155]$ & NA \\
\bottomrule
\end{tabular}
\end{table}

\begin{table}[ht!]
\caption{Using online assignment and estimated propensity scores. No pooling.} \label{tab:set2_online_estprop_nopool}
\centering \small
\begin{tabular}[t]{lllll}
\toprule
  & AIPW & AIPWl & IPW & Model-Based\\
\midrule
Point Estimate & 0.538 & 0.422 & 0.519 & 0.41\\
Gains & 0.197 & 0.08 & 0.178 & 0.068\\
Var(Gains) & (0.0097) & (0.0013) & (0.0072) & NA \\
CI of Gains & $[0.004, \; 0.390]$ & $[0.010, \; 0.150]$ & $[0.012, \; 0.344]$ & NA \\
\bottomrule
\end{tabular}
\end{table}

\end{document}